\documentclass[letterpaper, 10 pt, conference]{ieeeconf}  %
\IEEEoverridecommandlockouts

\usepackage{multirow}
\usepackage{cite} 
\usepackage[pdftex]{graphicx}
\usepackage[autostyle]{csquotes}
\usepackage{subcaption}
\usepackage{diagbox}
\usepackage[cmex10]{amsmath} 
\usepackage{amssymb}
\usepackage{booktabs} 

\makeatletter
\let\NAT@parse\undefined
\makeatother
\usepackage[pagebackref, colorlinks=true]{hyperref}

\title{\LARGE \bf
Multi-Task Learning for Automotive Foggy Scene Understanding via Domain Adaptation to an Illumination-Invariant Representation}

\author{Naif Alshammari$^{1,2}$, 
        Samet Ak\c{c}ay$^{1}$ and 
        Toby P. Breckon$^{1,3}$
\thanks{$^{1}$
        Department of Computer Science,
        Durham University, Durham, UK.}%
\thanks{$^{2}$
        Department of Natural and Applied Science,
        Majmaah University, Majmaah, KSA.}%
\thanks{$^{3}$
        Department of Engineering,
        Durham University, Durham, UK.}%
}

\begin{document}

\maketitle
\thispagestyle{empty}
\pagestyle{empty}

\begin{abstract}
Joint scene understanding and segmentation for automotive applications is a challenging problem in two key aspects:- (1) classifying every pixel in the entire scene and (2) performing this task under unstable weather and illumination changes (e.g. foggy weather), which results in poor outdoor scene visibility. These poor outdoor scene visibility leads to a non-optimal performance of deep convolutional neural network-based scene understanding and segmentation. In this paper, we propose an efficient end-to-end contemporary automotive semantic scene understanding approach under foggy weather conditions, employing domain adaptation and illumination-invariant image per-transformation. As a multi-task pipeline, our proposed model provides:- (1) transferring images from extreme to clear-weather condition using domain transfer approach and (2) semantically segmenting a scene using a competitive encoder-decoder convolutional neural network (CNN) with dense connectivity, skip connections and fusion-based techniques. We evaluate our approach on challenging foggy datasets, including synthetic dataset (\textit{Foggy Cityscapes}) as well as real-world datasets (\textit{Foggy Zurich} and \textit{Foggy Driving}). By incorporating RGB, depth, and illumination-invariant information, our approach outperforms the state-of-the-art within automotive scene understanding, under foggy weather condition. 
\end{abstract}

\IEEEpeerreviewmaketitle

\section{Introduction}
Scene understanding and pixel-wise segmentation is an active research topic requiring robust image pixel classification. However, the performance of the many state-of-the-art scene understanding algorithms is limited to clear weather conditions such that extreme weather and illumination variation could lead to inaccurate scene classification and segmentation \cite{finlayson2009entropy, maddern2014illumination, alvarez2011road, upcroft2014lighting, krajnik2015visual, corke2013dealing}. Up to now, too little attention has been paid to address the issue of automotive scene understanding under extreme weather conditions (foggy weather conditions as an example) \cite{foggy18, foggy19}, by proposing deep learning approaches generally applicable to ideal weather conditions only. This paper introduces an efficient algorithm that tackles the challenge of applicability of extreme weather conditions via a novel multi-task learning approach that translates foggy scene images to clear the scene and utilizes depth and luminance images for a superior semantic segmentation performance. 

\begin{figure}[t!]

	\begin{minipage}[b]{1.0\linewidth}
		\centering
		\centerline{\includegraphics{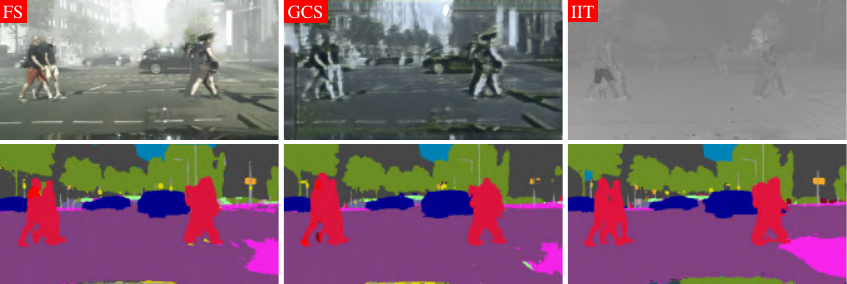}}

	\end{minipage}
	\vspace{-.55cm}
	\caption{Exemplar prediction results of the proposed approach.
	\textbf{FS}: Foggy Scene \cite{foggy18}; \textbf{GCS}: Generated Clear Scene using \cite{CycleGAN2017}; \textbf{IIT}: Illumination-Invariant Transform using \cite{maddern2014illumination}; along with the corresponding semantic segmentation outputs (ours).}
	\label{fig:intro}
	\vspace{-20px}
\end{figure}

\begin{figure*}[t!]
    \begin{minipage}[b]{1.0\linewidth}
        \centering
        \centerline{\includegraphics[width=\columnwidth]{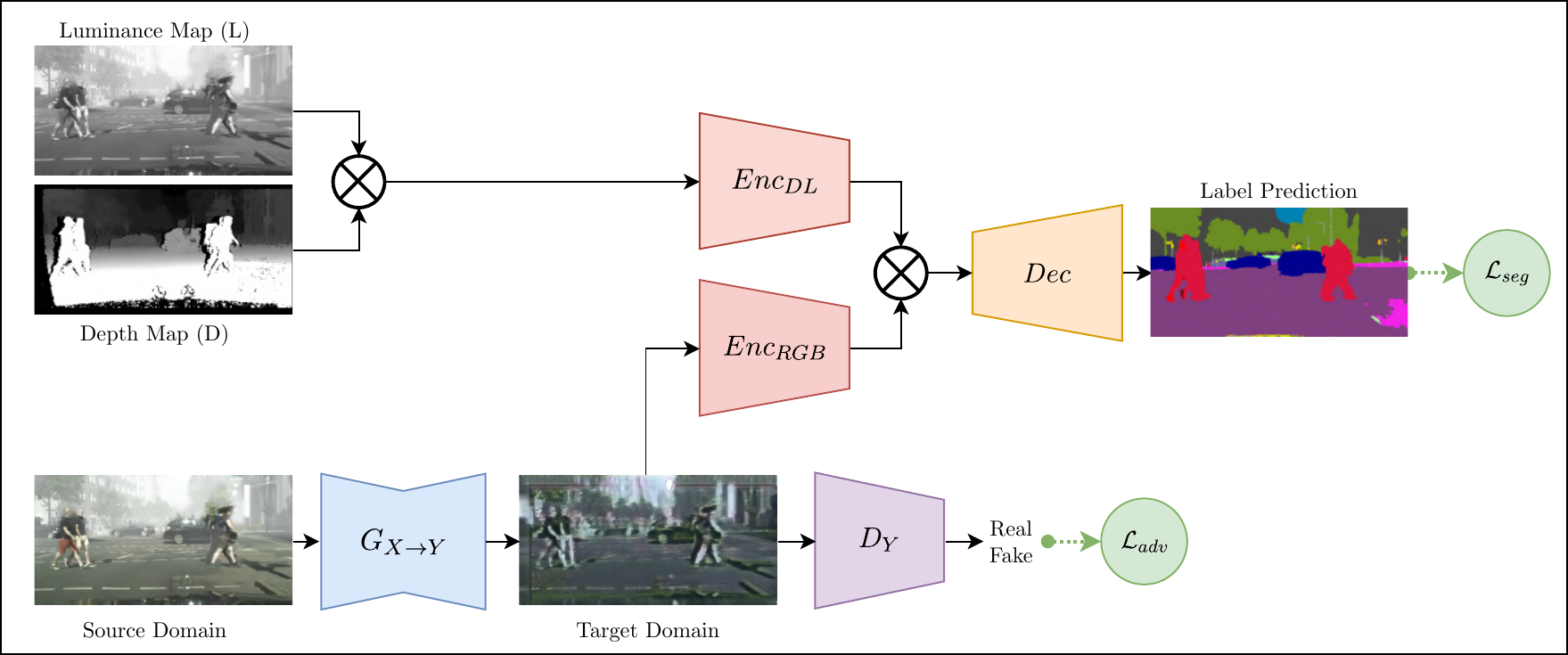}} 
    \end{minipage}
    \caption{Overview of our approach using \cite{CycleGAN2017, ldfnet}. The source domain $X$ (foggy scene) mapped to the target domain $Y$ (fine scene). Subsequently, the generated ($Y$) is fed to the RGB encoder in the semantic segmentation network. Depth (D) and luminance (L) image are incorporating RGB via DL encoder. Ultimately, the output from the two encoders is passed to the semantic segmentation decoder.}
    \label{fig:arch}
    \vspace{-20px}
\end{figure*}

Previous attempts to tackle the issue of scene understanding under non-ideal weather conditions for shadow removal and illumination reduction \cite{maddern2014illumination, alvarez2011road, upcroft2014lighting, krajnik2015visual}, haze removal and scene defogging \cite{defog1, defog2, defog3}, and foggy scene understanding \cite{foggy18, foggy19} are mostly based on conventional methods. Despite the general trend of the performance improvement within the automotive context \cite{pyramid, deeplab, resnet, refineNet17}, there is still room for further improvement via using more advanced deep neural networks. In parallel with using the recent techniques used in image segmentation \cite{densenet, fusenet, skip3, ldfnet}, employing the concepts of image-to-image translation to map from one domain to another \cite{CycleGAN2017, pix2pix2016} could be a useful step to colour transform that enables accurate semantic segmentation performance under extreme weather conditions.

In this work, we propose an efficient end-to-end automotive semantic scene understanding via multi-task approach, comprising both domain adaptation and segmentation network architectures together. These architectures are capable of benefiting from domain transformation, illumination-invariant image pre-transformation, depth and luminance information to achieve superior scene understanding and segmentation performance. As an effective technique to avoid information loss and share high-resolution features in the latter reconstruction stages during up-sampling of a CNN, we use skip-connections \cite{unet, skip1, skip2, skip3}. In addition, we use a fusion-based idea as an integration method within the overall model construction phases. To assess the impact on semantic segmentation performance, extensive experiments are also conducted using different invariant transformations (initial pre-process). 
\vspace{-7px}
\section{Related Work}
\label{sec:relatedwork}
Literature review is organized into three main categories: (1) semantic segmentation (Section \ref{sec:seg}), (2) domain transfer (Section \ref{sec:style}), and (3) illumination-invariant and perceptual colour space computation (Section \ref{sec:illum}).

\subsection{Semantic Segmentation}
\label{sec:seg}
Modern segmentation techniques utilize deep convolutional neural networks and outperform the traditional approaches by a large margin \cite{segnet17, deeplab, resnet, pyramid, refineNet17}. These contributions use a large dataset such as ImageNet \cite{imagenet} for pre-trained models. Recent segmentation techniques have their distinct characteristics by their designs such as:-- (1) the network topology for instance using: pooling indices \cite{segnet17}, skip connection \cite{unet}, multi-path refinement \cite{refineNet17}, pyramid pooling \cite{pyramid}, fusion-based architecture \cite{fusenet} and dense connectivity \cite{densenet}, (2) using alternative inputs such as depth as an extra channel RGB-D for the input image \cite{fusenet, holder_depth}, incorporate depth and luminance \cite{ldfnet}, and illumination invariant \cite{naif}, (3) inclusive or not adverse-weather conditions \cite{foggy18, foggy19}. As the main objective of this work is semantic segmentation under foggy weather condition, recent studies in this latter domain within the literature are specifically presented. \vspace{-.1cm}

\vspace{.2cm}
\noindent\textbf{{Foggy Scene Segmentation:}} 
Although studies have recognized the issue of foggy weather within scene understanding, there is limited research within the literature. One of these approaches, named SFSU \cite{foggy18}, shows a semi-supervised approach adapting methods in \cite{refineNet17, foggy18_model2} to perform scene understanding under foggy scenes using synthetic data. Generating Foggy Cityscapes \cite{foggy18} (partially synthetic dataset discussed in \ref{sec:foggy_city}) by adding fog to real images in the known datasets Cityscapes \cite{cityscapes}, the approach of \cite{foggy18} overcome the high cost of gathering and annotating data under extreme weather conditions. The above-mentioned supervised step is conducted to improve the semantic segmentation performance. Subsequently, this supervised learning combined with the unsupervised technique by augmenting clear weather images to their synthetic fog images. In another study, CMAda \cite{foggy19} proposes an adaptive semantic segmentation model from light synthetic fog to real dense fog. Developing a fog simulator, CMAda generates foggy datasets by adding synthetic fog into real images. Like \cite{foggy18}, CMAda  \cite{foggy19} is based on the RefineNet architecture \cite{refineNet17} as semantic segmentation.

\subsection{Domain Transfer}
\label{sec:style}
Transferring an image from its real domain to another differing domain allows multiple uses of such images taken in complex environments or generated in different forms. Using recent advances in the field of image style transfer, \cite{style_first}, where they generate the target image by capturing the style texture information of the input image by utilizing the Gram matrix. Work by \cite{amir_refer} shows that image style transfer (from the source domain to the target domain) is the process of minimizing the differences between source and target distribution. 
Recent methods \cite{pix2pix2016, CycleGAN2017, sketch} used Generative Adversarial Network (GAN) \cite{gan} to learn the mapping from source to target images. Based on training over a large dataset for specific image style,  \cite{CycleGAN2017} shows an efficient approach to transfer image style from one image into another.  Within the context of our work, we take advantage of \cite{CycleGAN2017} to improve the semantic segmentation by generating target scenes (clear-weather scenes) from the source domain (foggy scenes images) as a source image $I_{x}$ mapped into a target domain $I_{y}$ - hence significantly increasing our available image data training resource.
\vspace{-3.129px}

\subsection{Illumination-Invariant Images} 
\label{sec:illum} 
An illumination-invariant image $\mathcal{I}$ is a single channel image calculated by combining the three RGB colour channels in the image ${I}_{RGB} \in \{I_R, I_G, I_B\}$ that removes (or minimizes) scene colour variations due to varying scene lighting conditions. Inspired by \cite{finlayson2009entropy}, there has been numerous illumination-invariant image representation techniques \cite{finlayson2009entropy, maddern2014illumination, alvarez2011road, upcroft2014lighting, krajnik2015visual, corke2013dealing, kim2017pca} proposed in the literature. Mainly used for shadow removal in the outdoor scenes, illumination-invariant pre-processing provides better scene classification and understanding through reducing the illumination variations \cite{naif}. In this work, we take advantage of the image illumination-invariant transformation proposed in \cite{maddern2014illumination}, to be used as an alternative input to CNN-based scene understanding.
\section{Proposed Approach}
\label{sec:method}
\begin{figure*}[t!]
    \begin{minipage}[b]{1.0\linewidth}
        \centering
        \centerline{\includegraphics[width=\columnwidth]{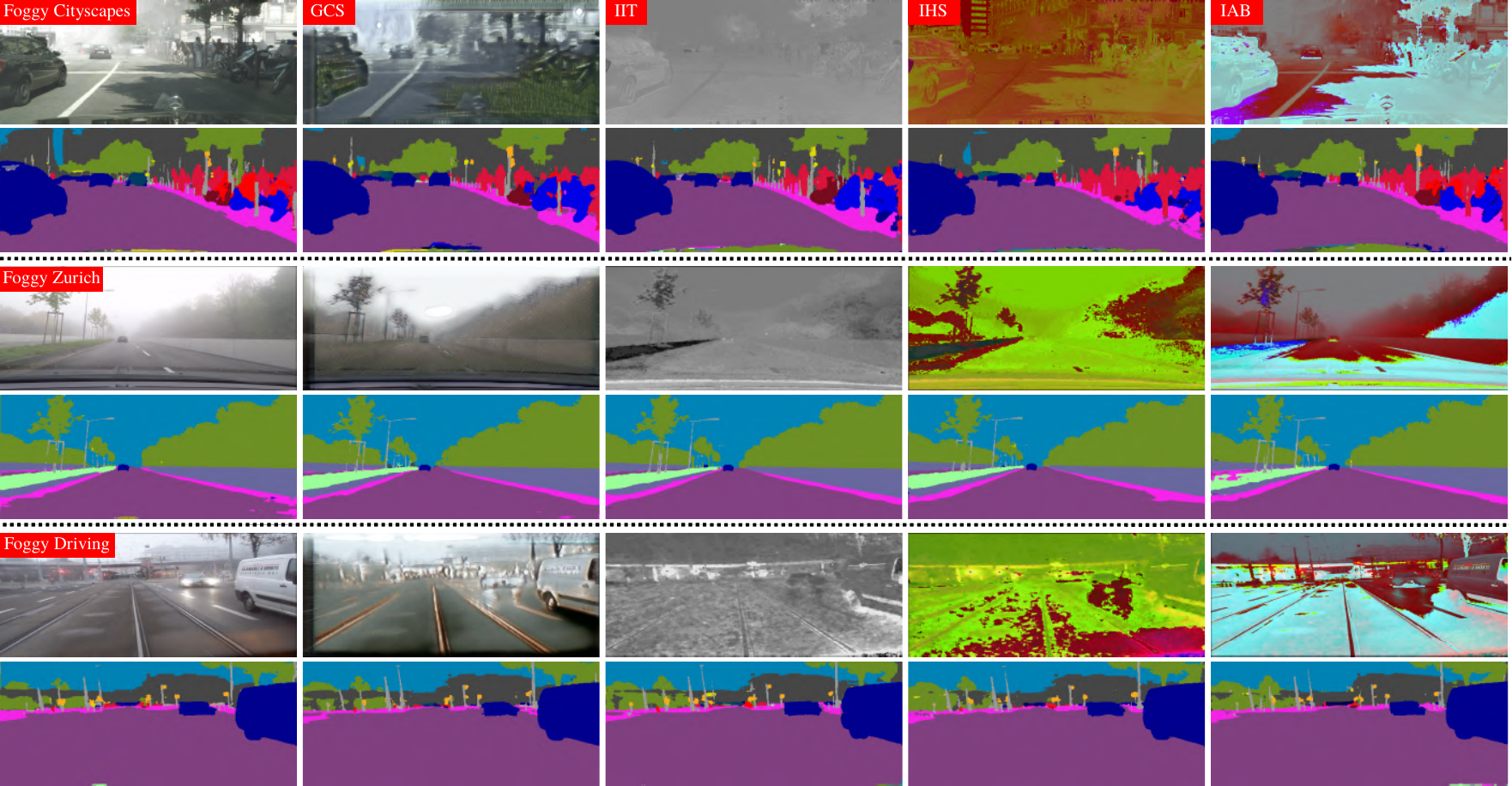}} 
    \end{minipage}
    \vspace{-.55cm}
    \caption{A comparison of semantic segmentation predictions on Cityscapes \cite{cityscapes} for the proposed approach. Left column shows results with no domain adaptation and image pre-transformation, followed by the four scenarios:- \textbf{GCS:} Generated Clear Scene \cite{CycleGAN2017}; \textbf{IIT:} Illumination-Invariant Transform \cite{maddern2014illumination}; \textbf{IHS:} Illumination combined HS in (HSV colour space); and \textbf{IAB:} Illumination combined with AB in (LAB colour space).}
    \label{fig:main_res}
    \vspace{-14px}
\end{figure*}

Our main objective is to train an end-to-end network that semantically labels every pixel in the scene that is invariant weather to conditions and illumination variations. In general, our approach consists of two sub-components, namely domain transfer and semantic segmentation (each of them could be functioning as an integrated unit). These sub-components produce two separate outputs: generated image (from fog to clear-weather) and semantic pixel labels. The pipeline of our approach is shown in Figure \ref{fig:main_res}. In this section, we will discuss the details of the two sub-components: Domain Transfer Model and Semantic Segmentation Model.\vspace{-3px} 


\subsection{Domain Transfer Model}
\label{sec:transfer}
Our goal is to learn mapping $\mathcal{D:} X \rightarrow Y$ from source domain $X$ (foggy scenes) to the target domain $Y$ (clear-weather) for which we assume such scene visibility level in the constructed image is the optimal input to the Semantic Segmentation Model (Section \ref{sec:seg}). We use generative adversarial networks proposed in \cite{CycleGAN2017} to generate the target images used later for the semantic segmentation task. A generator $G_{X \rightarrow Y}$ (generating clear scenes samples $Y'$) and a discriminator $D_{Y}$ (to discriminate between $Y$ and $Y'$) are used to perform the mapping function from the source and target domains. The loss for each generator $G$ with $D$ is calculated as follows: 

\begin{equation}\label{eq:adv_xy}
\begin{aligned}
\mathcal{L}_{adv}(X \rightarrow Y) = \min_{G_{Y \rightarrow X}} \max_{D_{Y}} \mathbb{E}_{y \sim \mathbb{P}_{d}(y)} [\log(D)_{(y)}]+ \\ \mathbb{E}_{x \sim \mathbb{P}_{d}(x)} [\log(1-D_{Y}(G_{X \rightarrow Y}(x)))]
\end{aligned}
\end{equation}where $\mathbb{P}_{d}$ is the data distribution, $X$ the source domain with samples $x$ and $Y$ the target domain with the samples $y$. 

\subsection{Semantic Segmentation Model}
\label{sec:semeseg} 
\begin{figure}
    \centering
    \includegraphics[width=\linewidth]{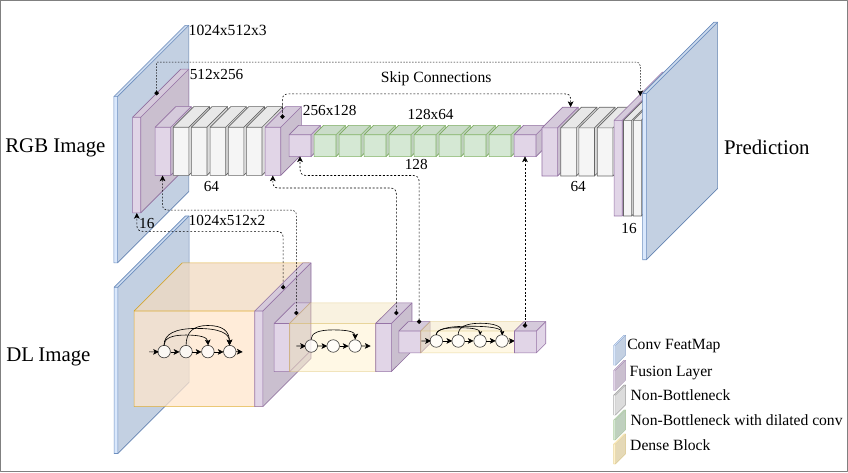}
    \caption{Details of the segmentation network which consists of two encoders taking two types of inputs: \textbf{RGB Image} and \textbf{DL Image} (with Depth and Luminance channels).}
    \vspace{-20px}
    \label{fig:seg-net}
\end{figure}

As a subsequent component to the overall model, our pipeline performs the task of semantic segmentation on generated images $Y'$ (target domain via $G_{X \rightarrow Y}(X)=Y'$) incorporated with depth $D$ and luminance $L$, via Semantic Segmentation Model) (shown in Fig. \ref{fig:seg-net}). Motivated by \cite{ldfnet}, we use an auto-encoder model for automotive semantic segmentation. The network design mainly consists of two encoders:- RGB encoder ($E_{RGB}$) and depth with luminance encoder ($E_{DL}$), to downsample the input image, and a decoder ($D$) to upsample the feature maps to the original input dimension. In addition, dense-connections and extracted fusion maps are implemented in the baseline architecture.

\vspace{.1cm}
\noindent\textbf{{RGB encoder:}}
\label{sec:rgb}
As designed to deal with a three-channel RGB input, RGB encoder ($E_{RGB}$) comprises three downsampler blocks with convolutional and max pooling layers followed by batch normalization and ReLu activation function (\{16, 64, 128\} respectively). Subsequently, five non-bottleneck modules are implemented including the factorized convolutions (convolution kernel $n$$\times$$n$ factorized into $n$$\times$$1$ and $1$$\times$$n$), each followed by batch normalization and ReLu with residual connections. With dilated and factorized convolutions, eight non-bottleneck modules were implemented as a last component of $E_{RGB}$. 

\vspace{.1cm}
\noindent\textbf{{DL encoder:}}
\label{sec:d&l}  
Unlike RGB encoder, depth and luminance encoder ($E_{DL}$) deals with depth and luminance images (concatenated as two-channel input). As a parallel functioning to $E_{RGB}$, $E_{DL}$ is designed with a dense connectivity technique for the information flow enhancement from earlier to last layers. Specifically, $E_{DL}$ consists of a downsampler (same as in $E_{RGB}$) followed by three dense blocks each has four, three, four modules respectively ($E_{DL}$ has the same number of channels as $E_{RGB}$). Each dense block is followed by a transition layer designed with $1$$\times$$1$ and followed with $2$$\times$$2$ average pool layer. As some datasets do not contain depth maps, we use a luminance only encoder ($E_{L}$) which is identical to ($E_{DL}$) except taking the only luminance channel.

$E_{RGB}$ and $E_{DL}$ are linked by fusing output layers from blocks sharing the same number of channels among $E_{RGB}$ and $E_{DL}$. The fusion connectivity is simply implemented by summing the two layers such that for inputs $x$ and $y$, the fused feature map is $E_{RGB}(x) + E_{DL}(y)$ or $E_{L}(y)$.

\vspace{.1cm}
\noindent\textbf{{Decoder:}}
\label{sec:dec}
After fusing the last extract feature maps from $E_{RGB}$ and either $E_{DL}$ or $E_{L}$, a decoder $D$ is performing upsample the feature maps to the original resolution. The upsampling is implemented in three stages \{64, 16, 19\}. In the first two stages, convolutional transpose, batch normalization and ReLu activation function, as well as two non-bottleneck modules, are employed. As a last component in the encoder, a convolutional transpose layer mapped the output to 19 class labels we aim to predict (class labels).

\vspace{.1cm}
Unlike LDFNet \cite{ldfnet}, we utilize skip connections for the fused features in the encoders into the decoder in order to avoid loss of the high-level spatial features before being downsampled. The fused feature maps \{64, 16\} passed from the encoders are concatenated with the corresponding upsampled feature maps in the decoder. As a semantic segmentation loss function, pixel-wise softmax with cross-entropy is used summing over all pixels within a patch as follows:

\begin{equation}\label{eq:seg1}
\begin{aligned}
{P}_{k}(x) =   \frac{e^{ak(x)}}{\sum_{k^\prime =1}^{K}e^{a_{k^\prime}(x)}},
\end{aligned}
\end{equation}

\begin{equation}\label{eq:seg2}
\begin{aligned}
\mathcal{L}_{seg} = -log(P_{l}(S(x))),
\end{aligned}
\end{equation}where $S(x)$ denotes to the output of the segmentation network, $a_{k}(x)$ the feature activation for the channel $k$, $K$ is the number of classes, $P_{k}(x)$ is the approximated maximum function and $l$ is the ground truth label. As a multi-task end-to-end pipeline, the joint loss function for the model with two tasks is calculated as follows:

\begin{equation}\label{eq:joint_loss}
    \mathcal{L} = \mathcal{L}_{adv}(I_{y}) +
                  \lambda_{seg}\mathcal{L}_{seg}(I_{s}),
\end{equation}where $\lambda_{seg}$ is a weighting coefficient, and empirically chosen as $0.10$ to balance the two losses.

\section{Dataset}
\label{sec:data}
The availability of numerous well-annotated datasets such as \cite {cityscapes, kitti, imagenet, pascal} has led to a proliferation in semantic segmentation studies. In this section, we will present the following datasets used in this paper:- \textit{Cityscapes dataset} \cite{cityscapes} as the base dataset representing clear scenes, \textit{Foggy Cityscapes dataset} \cite{foggy18} as a partially synthetic dataset where the fog is added into the clear-scenes, \textit{Foggy Driving} \cite{foggy18} and \textit{Foggy Zurich} \cite{foggy19} as real-world images where the adverse-weather (fog) is present. Besides, we will show the illumination-invariant pre-transformation used in our approach, as a technique to reduce the impact of such varying illumination conditions.

\begin{figure}[t!]
	\begin{minipage}[b]{1.0\linewidth}
		\centering
		\centerline{\includegraphics{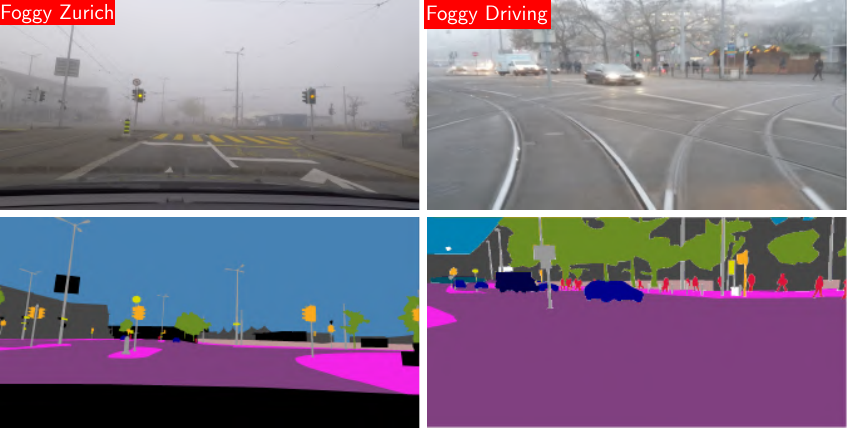}} 
	\end{minipage}
	\vspace{-.55cm}
	\caption{Sample images from \textbf{Foggy Zurich} \cite{foggy19} and \textbf{Foggy Driving} \cite{foggy18} along with their annotations.}
	\label{fig:zurich_imgs}
	\vspace{-14px}
\end{figure}

\vspace{.1cm}
\noindent\textbf{{Cityscapes Dataset:}}
We evaluate our approach on the Cityscapes \cite{cityscapes}, a large dataset for semantic segmentation of urban scenes. The dataset comprises of $2,975$ training and $500$ testing image examples (at a resolution of $1024\times2048$) with 19 pixel classes: \textit{\{road, sidewalk, building, wall, fence, pole, traffic light, traffic sign, vegetation, terrain, sky, person, rider, car, truck, bus, train, motorcycle and bicycle\}}. 

\vspace{.1cm}
\noindent\textbf{{Foggy Cityscapes Dataset:}}
\label{sec:foggy_city}
Foggy Cityscapes \cite{foggy18} is a partially synthetic dataset generated from real scenes from Cityscapes \cite{cityscapes} by adding synthetic fog to the real images using fog simulation \cite{foggy18} (the real images are taken in the clear-weather conditions). Three different versions of this dataset depending on fog density level (controlled using attenuation coefficient $\beta \in \{0.005, 0.01, 0.02\}$ - from light to dense fog) are used. As real scenes dataset is provided with fine annotated images, we employ the annotations as labels for the synthetic foggy datasets. Foggy Cityscapes dataset consists of $8925$ training and $1500$ testing image examples (resolution on $1024\times2048$).

\begin{table*}[t]
	\centering
	\begin{tabular}{c|c|c|c|c|c|c|c|c|c}
		Dataset & \multicolumn{3}{c|}{Foggy Zurich} & \multicolumn{3}{c|}{Foggy Driving} & \multicolumn{3}{c}{Foggy Cityscapes} \\ \hline \hline
		\diagbox{Methods}{Results} & {Global avg.} & {Class avg.} & {Mean IoU} & {Global avg.} & {Class avg.} & {Mean IoU} & {Global avg.} & {Class avg.} & {Mean IoU} \\ \hline \hline
		AdSegNet \cite{similar_naif}      & \multicolumn{2}{c|}{\textemdash}   & 0.25 & \multicolumn{2}{c|}{\textemdash} & 0.44 & \multicolumn{3}{c}{\textemdash}               \\ \hline
		SFSU \cite{foggy18}      & \multicolumn{2}{c|}{\textemdash}   & 0.35 & \multicolumn{2}{c|}{\textemdash} & 0.46 & \multicolumn{3}{c}{\textemdash}               \\ \hline
		CMAda2+ \cite{foggy_old}      &\multicolumn{2}{c|}{\textemdash}   & 0.43 & \multicolumn{2}{c|}{\textemdash} & 0.49  & \multicolumn{3}{c}{\textemdash}              \\ \hline
		CMAda3+ \cite{foggy19}      &\multicolumn{2}{c|}{\textemdash}  & 0.46 & \multicolumn{2}{c|}{\textemdash} & 0.49  & \multicolumn{3}{c}{\textemdash}             \\ \hline
		\multicolumn{10}{c}{Our approach} \\ \hline
		IAB         & 0.84                & 0.54               & 0.43   & 0.91                & 0.65               & 0.54 & 0.89                & 0.69               & 0.54            \\ \hline
		IHS        & 0.89                & 0.61               & 0.45 & 0.90                & 0.62               & 0.51 & 0.88                & 0.66               & 0.53           \\ \hline
		FS    & 0.89                   & 0.60               &  0.51 & 0.88 & 0.58 & 0.44 & 0.90                   & 0.69               &  0.56               \\ \hline		
		IIT     & 0.91                   &\textbf{0.63}               & 0.52 & 0.91                   & 0.67               & \textbf{0.54} & \textbf{0.92}                   & 0.70               & \textbf{0.59}               \\ \hline
		GCS     & \textbf{0.94}                   & 0.60              & \textbf{0.54} & 0.89 & \textbf{0.72} & \textbf{0.59} & \textbf{0.92}                   & \textbf{0.71}               & \textbf{0.60}              \\ \hline
	\end{tabular}
	\vspace{3px}
	\caption{Quantitative comparison of semantic segmentation over the Foggy Zurich \cite{foggy19} and Foggy Driving \cite{foggy18} datasets of scenes with 19 classes. Our approach addressed five variants: with pre-transform as Illumination-Invariant Transform \textbf{IIT} \cite{maddern2014illumination}, combined with ${AB} \in LAB$ as \textbf{IAB}, or with $HS \in HSV$ as \textbf{IHS}; with no transform (Foggy Scenes) \textbf{FS} in the presence of fog; and with Generated Clear Scene \textbf{GCS} using \cite{CycleGAN2017}.}
	\label{tab:main_res}
	\vspace{-14px}
\end{table*}

\vspace{.1cm}
\noindent\textbf{{Foggy Driving Dataset:}}
Foggy Driving dataset \cite{foggy18} (Fig. \ref{fig:zurich_imgs}) is a real-world dataset collected in the foggy-weather condition, consisting of $101$ images (at a resolution of $960\times1280$) with annotations for semantic segmentation and object detection tasks. Following Cityscapes dataset, Foggy Driving dataset is labelled with 19 classes.  

\vspace{.1cm}
\noindent\textbf{{Foggy Zurich Dataset:}}
Foggy Zurich \cite{foggy19}  (Fig. \ref{fig:zurich_imgs}) is a real foggy-scenes dataset consisting of $3808$ images (at the resolution of $1920\times1080$) collected in the city of Zurich. Using Cityscapes approach, Foggy Zurich provides pixel-level annotations for 40 scene, including dense fog. 

\vspace{.1cm}
\noindent\textbf{{Illumination-Invariant Pre-transformation:}}
The illumination-invariant image, where global illumination variation and localised shadows are significantly reduced within the scenes, is computed using the approach proposed in \cite{maddern2014illumination}. To generate such an invariant representation, a 3- channel floating point RGB image (${I}_{RGB} \in \{I_R, I_G, I_B\}$) converted into the corresponding illumination-invariant image as follows. 

\begin{equation}\label{eq:maddern}
    \begin{aligned} 
        {\mathcal{I}}_{\text{Maddern}}=\quad &0.5+\log(I_G)-\alpha log(I_B)-\\
        &(1-\alpha) \log(I_R),
    \end{aligned} 
\end{equation}where $\alpha = 0.48$ for the reference camera in use (Point Grey Bumblebee-2), and $0.5$ for pixels normalised into the range $\{0 ... 1\}$. For further evaluation, we use the illumination-invariant channel combined with ${AB} \in LAB$ as \textbf{IAB}, and with $HS \in HSV$ as \textbf{IHS} to assess their impact on the performance of our model.

\vspace{.1cm}
\noindent\textbf{{Luminance Transformation:}}
Luminance transformation is a translated grayscale image $L$ generated from ${I}_{RGB} \in \{I_R, I_G, I_B\}$ to reduce the noise and improve getting better feature extraction, defined as follows: 

\begin{equation}\label{eq:Limg}
    \begin{aligned} 
        L=0.299(I_R) + 0.587(I_G) + 0.144(I_B)
    \end{aligned} 
\end{equation}

\section{Implementation Details}
\label{sec:implement}
We implement our approach in PyTorch \cite{pytorch}. For optimization, we employ ADAM \cite{adam} with an initial learning rate of $5\times10^{-3}$ and momentum of $\beta_1 = 0.5, \beta_2 = 0,999$. The weighting coefficient in the loss function is empirically chosen to be $\lambda_{seg} = 0.10$. By following \cite{enet} and \cite{ldfnet}, we weight the classes of the dataset duo to imbalance number of pixels of each class in the dataset as follows:

\begin{equation}\label{eq:cweight}
	\begin{aligned} 
		\omega_{class}=\frac{1}{ln(c+p_{class})}
	\end{aligned}
	\vspace{0.2cm} 
\end{equation} where c is an additional parameter set to 1.10 to restrict the class weight and $p_{class}$ is the probability of belonging to that class. We train the model for $100$ epoch by using NVIDIA Titan X and GTX 1080Ti GPUs. Like \cite{ldfnet}, we apply data augmentation in training using random horizontal flip. For semantic accuracy evaluation, we use the following evaluation measures:-
(1) class average accuracy, the mean of the predictive accuracy over all classes, (2) global accuracy, which measures overall scene pixel classification accuracy, and (3) mean intersection over union (mIoU). 



\vspace{.1cm}
\section{Evaluation}
We evaluate the performance of automotive scene understanding and segmentation using the modified LDFNet \cite{ldfnet} CNN architecture, on the  \textit{Cityscapes} \cite{cityscapes}, \textit{Foggy Driving} \cite{foggy18},  and  \textit{Foggy Zurich} \cite{foggy19} datasets. The evaluation was performed as follows:

\vspace{.1cm}
\renewcommand{\labelenumii}{\Roman{enumii}}
\begin{enumerate}
    \item \label{point1} we train the semantic segmentation model (Fig. \ref{fig:seg-net}) (employed later as a sub-model in the overall model (Fig. \ref{fig:arch}) trained in step \ref{point2}) on the Foggy Cityscapes (partially synthetic datasets).
     \item \label{point1_1} we fine-tune the model trained in step \ref{point1} on Foggy Zurich and Foggy Driving dataset (real-world datasets).
    \item \label{point2} foggy datasets used in the model trained in step \ref{point1} are first mapped to clear scenes using the domain adaptation sub-model (Fig. \ref{fig:arch}), and subsequently the generated images are fed into the semantic segmentation sub-model (Fig. \ref{fig:arch}) to train the second-task (semantic segmentation). 
    \item \label{point3} we fine-tune the model obtained from step \ref{point2} on Foggy Zurich and Foggy Driving dataset.
    \item we generate the illumination-invariant transform IIT and perceptual colour-space IAB and IHS from the foggy datasets and use them as alternative inputs for the model trained in step \ref{point1}, to assess their impact.
\end{enumerate}

\begin{table*}[th]
    \resizebox{\linewidth}{!}{
        \small{
            \begin{tabular}{c|c|c|c|c|c|c|c|c|c|c|c|c|c|c|c|c|c|c|c}
                
                \multicolumn{1}{c}{Method}                   & \multicolumn{1}{c}{\rotatebox{90}{Road}} & \multicolumn{1}{c}{\rotatebox{90}{Sidewalk}} & \multicolumn{1}{c}{\rotatebox{90}{Building}}  & \multicolumn{1}{c}{\rotatebox{90}{Fence}}  & \multicolumn{1}{c}{\rotatebox{90}{Wall}} & \multicolumn{1}{c}{\rotatebox{90}{Vegetation}} & \multicolumn{1}{c}{\rotatebox{90}{Terrain}} & \multicolumn{1}{c}{\rotatebox{90}{Car}} & \multicolumn{1}{c}{\rotatebox{90}{Truck}} & \multicolumn{1}{c}{\rotatebox{90}{Train}} & \multicolumn{1}{c}{\rotatebox{90}{Bus}} & \multicolumn{1}{c}{\rotatebox{90}{Bicycle}} & \multicolumn{1}{c}{\rotatebox{90}{Motorcycle}} & \multicolumn{1}{c}{\rotatebox{90}{Sky}} & \multicolumn{1}{c}{\rotatebox{90}{Pole}} & \multicolumn{1}{c}{\rotatebox{90}{Traffic-sign}} & \multicolumn{1}{c}{\rotatebox{90}{Traffic-light}} & \multicolumn{1}{c}{\rotatebox{90}{Person}} & \multicolumn{1}{c}{\rotatebox{90}{Rider}}\\ \hline \hline
                
                FS & 95.1     & 65.8 & 78.8 & 29.1 & 39.6        & 49.4 & 45.6       & 53.0  & 85.7        & 59.9     & 70.0      & 58.6       & 26.8  & 81.4 & 51.3 & 36.2 & 02.1 & 42.8 & 47.0      \\ \hline
                IAB & 94.6 & 71.0 & 78.4 & 31.5 & 43.2 & 47.4 & 47.3 & 54.2 & 85.7 & 58.3 & 64.4 & 51.2 & 34.4 & 71.3 & 41.4 & 41.1 & 22.8 & 48.0 & 45.5  \\ \hline         
                IHS & 95.2 & 71.7 & 77.5 & 29.8 & 35.6 & 44.5 & 49.6 & 54.8 & 84.7 & \textbf{63.2} & 63.8 & 49.8 & 28.4 &72.5 & 30.2 & 41.0 & 16.3 & \textbf{57.5} & 50.4  \\ \hline
                IIT   & \textbf{95.6} & \textbf{74.2} & 80.2 & \textbf{34.9} & 44.5 & \textbf{52.2} & 49.4 & \textbf{57.6} & 86.6 & \textbf{61.3} & 67.2 & \textbf{66.5} & \textbf{53.3} & 80.5 & \textbf{60.1} & 48.2 & 27.5 & \textbf{50.7} & \textbf{53.4}   \\ \hline
                GFS     & 95.0 & 66.4 & \textbf{84.4} & 24.3 & \textbf{45.8} & 51.6 & \textbf{50.9} & 56.3 & \textbf{87.9} & 59.9 & \textbf{84.7} & 63.6 & 45.6 & \textbf{85.0} & 55.8 & 47.7 & \textbf{33.8} & 50.2 & 49.2 \\ \hline

            \end{tabular}
    }}
    \vspace{1px}
    \caption[Cityscapes quantitative results.]{Class IoU results on the Foggy Cityscapes \cite{foggy18} using five evaluation methods. \textbf{FS:} Foggy Scene (no transform); \textbf{IAB:} illumination-invariant compined with AB in (LAB); \textbf{IHS:} Illumination-Invariant combined with HS in (HSV); \textbf{GCS:} Generated Clear Scene \cite{CycleGAN2017}; \textbf{IIT:} Illumination-Invariant Transform using \cite{maddern2014illumination}.}
    \label{tab:Cityscapes_ciou}
\end{table*}

We will present and analyze the results of each step above in the next of this section.

\vspace{.1cm}
\noindent\textbf{{Foggy Scenes (FS):}} As an initial stage, we evaluate the performance of semantic segmentation on foggy scene datasets: \{\textit{Foggy Cityscapes} \cite{foggy18}, \textit{Foggy Driving} \cite{foggy18}, and \textit{Foggy Zurich} \cite{foggy19}\}. Here, we deal with the semantic segmentation model (shown in Fig \ref{fig:seg-net}) as an independent model and isolated from the whole pipeline to investigate its performance on the foggy dataset. With an improvement of ($5\%$), our approach achieves the heights mean intersection-over-union (mIoU) accuracy ($51\%$) when compared with the reference results of CMAda \cite{foggy19} on \textit{Foggy Zurich} \cite{foggy19}. However, the reference results of CMAda \cite{foggy19} produce the highest mIoU ($49\%$) on \textit{Foggy Driving} \cite{foggy18} (see Table \ref{tab:main_res}). As the scene visibility is significantly poor in the foggy weather condition, this method produces the lowest accuracy (when compared with using domain adaptation and illumination-invariant pre-transformation techniques discussed later) in the evaluation measures: (overall, class average, and mean IoU accuracy) ($90\%, 69\%, 56\%$) respectively using (\textit{Foggy Cityscapes}), ($88\%, 58\%, 44\%$) (\textit{Foggy Driving}), and ($89\%, 60\%, 51\%$) (\textit{Foggy Zurich}) (see Table \ref{tab:main_res}). Within individual class performance, this method fails to achieve any improvement (Table \ref{tab:Cityscapes_ciou}) when compared with the methods discussed in the next of this section.


\vspace{.1cm}
\noindent\textbf{{Generated Clear Scenes (GCS):}}
The second evaluation is performed on the generated datasets (with clear scenes) using domain transfer approach shown in Section \ref{sec:transfer}. We mapped the source domain (foggy scenes) in the aforementioned foggy datasets into the target domain (clear scenes), assuming that will increase the level of visibility and lead to better scene understanding and segmentation. In this stage, we make use of the two models (domain transfer and semantic segmentation) simultaneously in one pipeline (Fig. \ref{fig:arch}). As a result, this method outperforms the reference results of \cite{foggy18, foggy19} and the above method on all the evaluation measures (overall, class average, mIoU) ($92\%, 71\%, 60\%$) respectively using \textit{Cityscapes}, ($89\%, 72\%, 59\%$) (\textit{Foggy Driving}), and ($94\%, 60\%, 54\%$) (\textit{Foggy Zurich})  (see Table \ref{tab:main_res}). With an improvement of ($8\%$) and ($10\%$) in mIoU on \textit{Foggy Zurich} and \textit{Foggy Driving} respectively, this method shows a large impact of employing domain transfer technique to map the complex weather condition to the clear one. For per-class performance, this method achieves higher results (on \textit{Foggy Cityscapes}) in (cIoU) (seven classes) among the five methods conducted in this work (see Table \ref{tab:Cityscapes_ciou}), placing it second when compared with the other methods in this manner. 


\vspace{.1cm}
\noindent\textbf{{Illumination-Invariant Transform (IIT):}} As a slight difference over the first evaluation,  we train the semantic segmentation model on the illumination-invariant images generated from the foggy datasets (\textit{Foggy Cityscapes}, \textit{Foggy Driving}, \textit{Foggy Zurich})  using the approach of \cite{maddern2014illumination}. In other words, we are replacing foggy dataset (RGB images) with illumination-invariant images to see their impact on the CNN-based model performance. This method achieves the highest mIoU accuracy ($52\%$) and ($54\%$) on (\textit{Foggy Zurich}) and (\textit{Foggy Driving}) respectively when compared with the reference results of \cite{foggy18, foggy19} (see Table \ref{tab:main_res}). However, this technique has achieved the second highest accuracy in (overall, class average, and mIoU accuracy) ($92\%, 70\%, 59\%$) respectively using \textit{Cityscapes}, ($91\%, 67\%, 54\%$) (\textit{Foggy Driving}), and ($91\%, 63\%, 52\%$) (\textit{Foggy Zurich}) when compared with the above methods (see Table \ref{tab:main_res}). As per-class comparison, superior performance in the majority of classes (eleven classes) (using \textit{Foggy Cityscapes}) among the five methods conducted in this work (see Table \ref{tab:Cityscapes_ciou}) has achieved using \textit, ranking it in the top.\vspace{.1cm} 


\section{Conclusion}
\label{sec:conclusion}
This paper proposes a novel end-to-end automotive semantic scene understanding via domain transfer to an illumination-invariant representation. The proposed model can semantically predict per pixel scene labels under extreme foggy-weather conditions. The use of domain adaptation maps a scene taken in foggy conditions to a target domain considered optimal (clear-weather) with visibility scene increased. As a result, the performance of deep convolutional network-based scene understanding and segmentation under weather-condition has progressively improved \cite{foggy18,foggy19}. As another way to improve scene understanding, we use an illumination-invariant pre-transformation technique, with and without hybrid colour information, applied as alternative inputs to semantic segmentation network. By examining these transforms, we are able to show that pre-processing can influence trained network performance. Using fusion-based architecture, dense connectivity and skip-connections for the feature fusion, our approach achieves significant results over the state-of-the-art semantic segmentation under foggy weather condition \cite{foggy18, foggy19, foggy_old}. \vspace{.1cm} 

\bibliographystyle{IEEEtran}
\bibliography{ref}

\end{document}